**Title: Deep Learning for Automated Classification of Tuberculosis-Related Chest X-Ray: Dataset Specificity Limits Diagnostic Performance Generalizability**

Authors: Seelwan Sathitratanacheewin (1 and 2) and Krit Pongpirul (1, 2, and 3)

((1) Department of Preventive and Social Medicine, Faculty of Medicine, Chulalongkorn University, Bangkok, Thailand, (2) Thai Health AI Foundation, Bangkok, Thailand, (3) Department of International Health and Department of Health, Behavior, and Society, Johns Hopkins Bloomberg School of Public Health, Baltimore, MD, USA)

Machine learning has been an emerging tool for various aspects of infectious diseases including tuberculosis surveillance and detection. However, WHO provided no recommendations on using computer-aided tuberculosis detection software because of the small number of studies, methodological limitations, and limited generalizability of the findings. To quantify the generalizability of the machine-learning model, we developed a Deep Convolutional Neural Network (DCNN) model using a TB-specific CXR dataset of one population (National Library of Medicine Shenzhen No.3 Hospital) and tested it with non-TB-specific CXR dataset of another population (National Institute of Health Clinical Centers). The findings suggested that a supervised deep learning model developed by using the training dataset from one population may not have the same diagnostic performance in another population. Technical specification of CXR images, disease severity distribution, overfitting, and overdiagnosis should be examined before implementation in other settings.

**Deep Learning for Automated Classification of Tuberculosis-Related Chest X-Ray: Dataset Specificity Limits Diagnostic Performance Generalizability**

**Background:**

Tuberculosis (TB) is a major health problem in many regions of the world, especially developing countries. As part of the World Health Organization (WHO) systematic screening strategy to ensure early and correct diagnosis for all people with TB, Chest X-ray (CXR) is one of the primary tools for triaging and screening for TB because of its high sensitivity, depending on how the CXR is interpreted [1]. However, significant intra- and inter-observer variations in the reading of CXR can lead to overdiagnosis or underdiagnosis of tuberculosis.

Deep convolutional neural network (DCNN) has emerged as an attractive technique for TB surveillance and detection. This 'supervised' machine learning algorithm learns a mapping from a set of covariates to the outcome of interest by using the training data then applies this mapping to the new test data for identification or prediction tasks [2]. In a common deep learning model, the covariates are the color pixel values of the CXR images whereas the outcome is the radiologist's interpretation and impression of the corresponding CXR.

Recently, the National Institute of Health (NIH) released the ChestX-ray8 dataset with more than 100,000 anonymized CXR images and their associated data which compiled from more than 32,000 patients [3]. These data allow researchers to further develop an algorithm for classifying lung abnormalities labeled from the radiological reports using National Language Processing technique [4, 5]. Nonetheless, according to the ChestX-ray8 criteria, a CXR image with minimal lung lesions could be incorrectly labeled as normal whereas these radiologic labels are not specific to TB [3].

The Computer Aided Detection for Tuberculosis (CAD4TB), SemanticMD, and Qure.ai are selected examples of currently available computer-aided detection (CAD) software. CAD4TB is TB specific and had demonstrated a good diagnostic performance [6, 7] but still inferior to that of expert readers [8]. As of 2016, WHO provided no recommendations on using CAD for TB because of small number of studies, methodological limitations, and limited generalizability of the findings [1]. The other two software cover a broader range of clinical conditions but available only for investigational use in the U.S.

Unlike other diagnostic tests, technical specification of CXR images and disease severity distribution could affect the diagnostic performance of a supervised machine learning model. That is, the picture archive and communication system (PACS) data has more information but requires more time, storage, and processing power than compressed picture formats whereas the supervised deep learning model developed based on CXR images of hospitalized patients would be smarter at detecting severe case than another community-based model. To quantify the extent of dataset specificity that limits the generalizability of CAD for TB, we developed a DCNN model using a TB-specific CXR dataset of one population and tested it with non-TB-specific CXR dataset of another population.

**Materials and Methods:**

Two de-identified HIPAA-compliant datasets, the National Library of Medicine (NLM) Shenzhen No.3 Hospital X-ray set and the NIH ChestX-ray8 database were included in this study. The Shenzhen dataset collected 336 normal and 326 abnormal CXR showing various manifestations of TB in JPEG format as part of the routine care at Shenzhen No.3 Hospital in Shenzhen, Guangdong providence, China. The ChestX-ray8 comprises CXR images acquired as a part of routine care at NIH Clinical Center, Bethesda, Maryland, USA. It comprises of 112,120 frontal view CXR images of 32,717 unique de-identified patients with the text mined eight disease image labels (each image can have more than one label), from their corresponding radiological reports using natural language processing.

Firstly, the Shenzhen dataset was split into training (75%), validation (15%), and intramural test (10%) sets. Based on the TensorFlow framework, Inception V3, a novel pre-trained DCCN, was augmented with several techniques to classify each image as having TB characteristics or as healthy. Next, 112,120 CXR (60,362 normal and 51,760 abnormal images with one of fourteenth common thoracic abnormalities including atelectasis, cardiomegaly, consolidation, edema, effusion, emphysema, fibrosis, hernia, infiltration, mass, nodule, pleural thickening, pneumonia and pneumothorax) from ChestX-ray8 dataset were used to construct an extramural test set to examine the generalizability of the DCNN model to classify normal and other CXR in addition to test set from intramural (Shenzhen) dataset. Lastly, the prevalence of TB-associated CXR in the ChestX-ray8 dataset was estimated by using the final DCNN model. Receiver operating characteristic (ROC) curves and areas under the curve (AUC) were used to assess model performance and to define the optimal cut point for TB detection.

**Results:**

In the training and intramural test sets using Shenzhen hospital database, the DCCN model exhibited an AUC of 0.9845 and 0.8502 for detecting TB, respectively (Figure 1, 2). However, the AUC of the supervised DCNN model in ChestX-ray8 dataset was dramatically dropped to 0.7054 (Figure 3). Using the cut points at 0.90 which suggested 72% sensitivity and 82% specificity in the Shenzhen dataset, the final DCNN model estimated that 36.51% of abnormal radiographs in the ChestX-ray8 dataset were related to TB.

Figure 1 ROC Curves of DAC4TB in the training Shenzhen No.3 Hospital set.

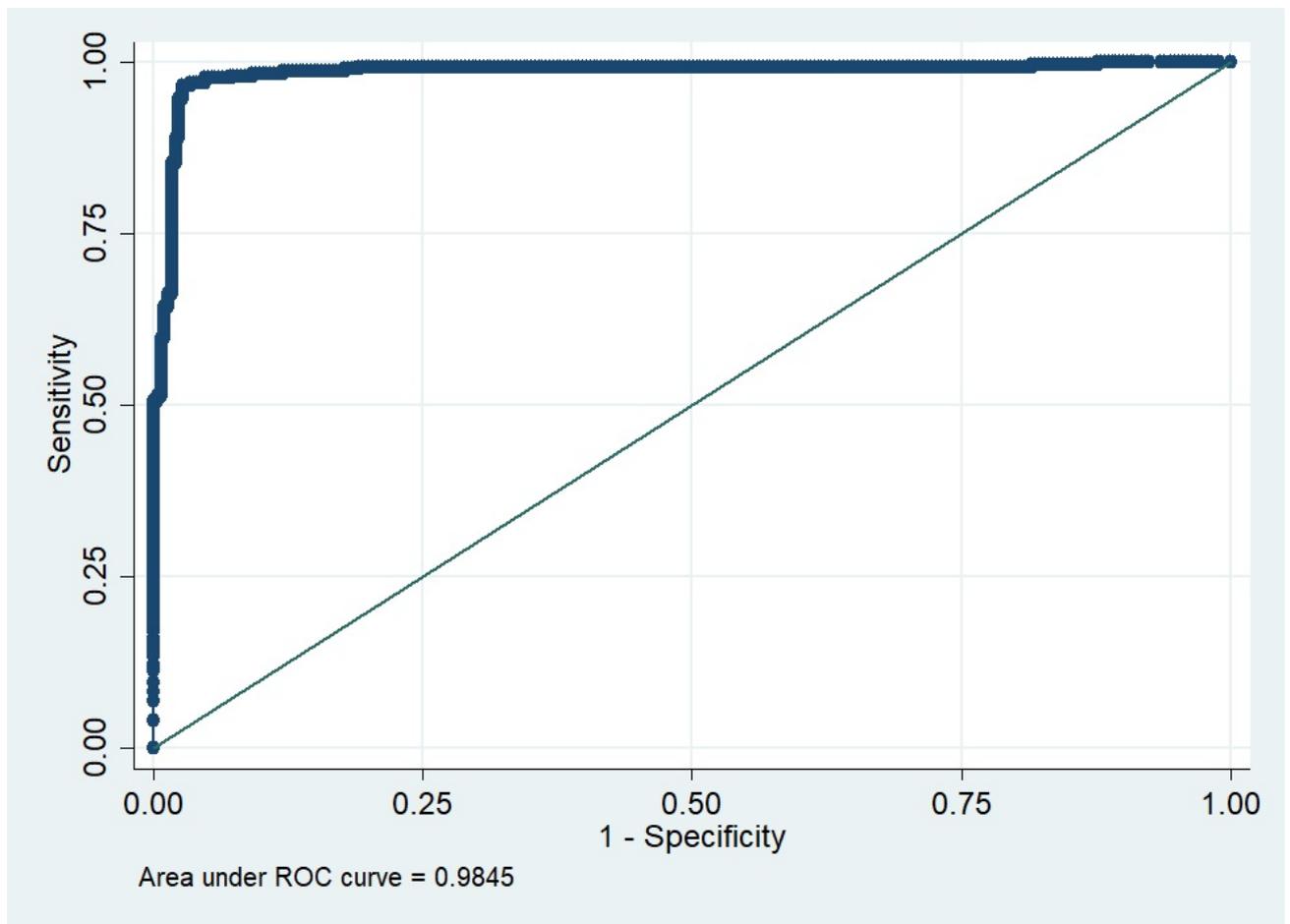

Figure 2 ROC Curves of DAC4TB in the intramural Shenzhen No.3 Hospital test set.

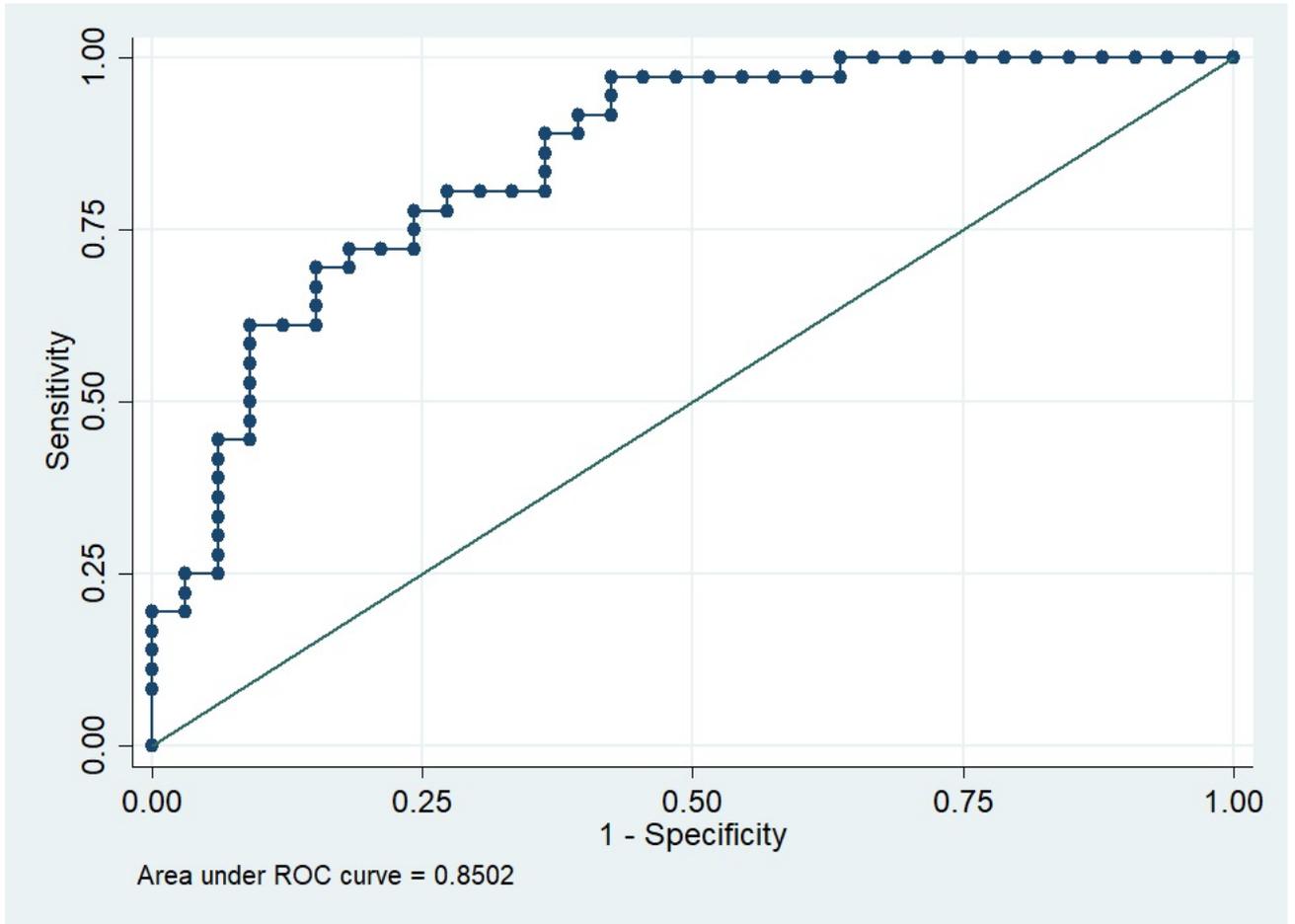

Figure 3 ROC Curves of DAC4TB in the extramural NIH ChestX-ray8 test set.

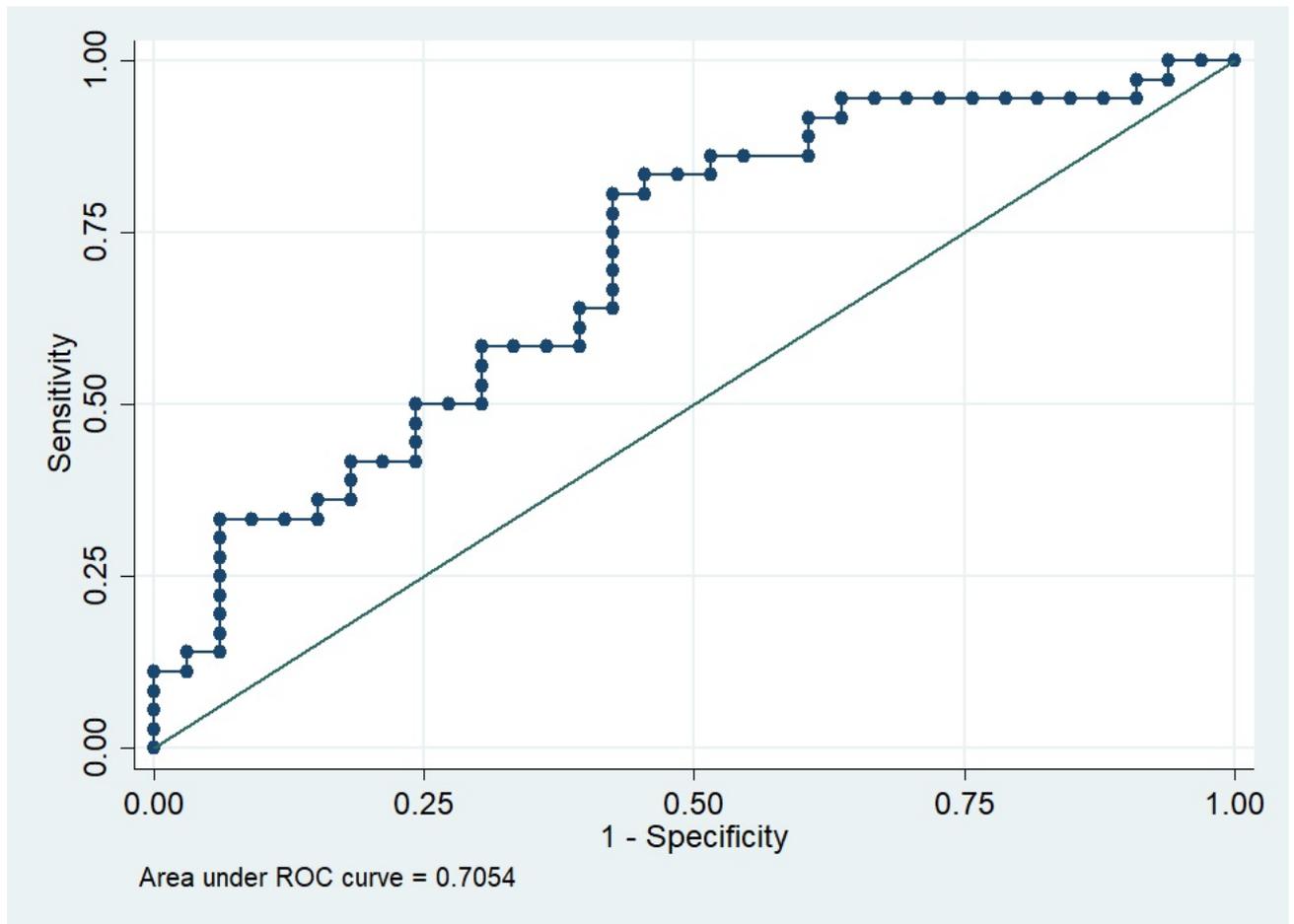

**Discussion:**

Our findings suggested that the diagnostic performance of a supervised machine learning model is dataset specific, because of varying technical specification of CXR images and disease severity distribution in different population. In this study, the Inception V3 Tensorflow model maintained the diagnostic accuracy for the training, validation, and intramural test images in Shenzhen dataset. This ability to correctly classified unseen test images within the same dataset is known as generalization. However, the significant drop in the diagnostic performance in the extramural test images reflects poor generalizability.

Falling in performance of the model on new dataset but well fitted the training dataset is called 'overfitting'. Also, this indicates the dataset specificity. One might argue that overfitting becomes more apparent as our training dataset is relatively small. However, variation in CXR readings between the two datasets as differences in prevalence of common lung abnormalities and radiologists practice pattern in China and U.S. population are more likely to play a role in overfitting in our case. Further researches need to examine these issues.

The 36.51% of abnormal CXR in the NIH ChestX-ray8 dataset was associated with TB might demonstrate the 'overdiagnosis' of deep learning since many CXR abnormalities that are compatible with pulmonary TB are seen also in several lung pathologies and, therefore, are indicative not only of TB but also of other pathologies.

Some limitations of this study should be noted. Like other deep learning studies, the images were resized to a manageable dimension before being fed into the model as a larger file will increase the training time and will require more robust central and graphical processing power. Accuracy may be improved by using higher resolution images, particularly for subtle findings, and more research is needed in this regard. Lastly, this retrospective study was conducted using the datasets that were available at the time. Further investigation on the use of deep learning in a real world large scale screening program for pulmonary TB in prevalent regions is essential.

**Conclusion:**

A supervised deep learning model developed by using the training dataset from one population may not always have the same diagnostic performance in another population. Technical specification of CXR images, disease severity distribution, overfitting, and overdiagnosis should be examined before implementation in other settings.